\pdfoutput=1

\documentclass[11pt]{article}

\usepackage[]{acl}
\usepackage{times}
\usepackage{latexsym}
\usepackage{makecell}
\usepackage{bm}
\usepackage{mathtools}
\usepackage{amsthm}
\usepackage{amsmath, extpfeil}
\usepackage[T1]{fontenc}

\usepackage[utf8]{inputenc}
\usepackage{subfigure}
\usepackage{microtype}
\usepackage{enumitem}
\usepackage{graphicx}
\usepackage{multirow}
\usepackage{amsmath}
\usepackage{amssymb}
\usepackage{algorithm}
\usepackage{algorithmicx}
\usepackage{algpseudocode}
\usepackage{xcolor}
\usepackage{array}
\usepackage{booktabs}
\usepackage{arydshln}

\title{Poor-Supervised Evaluation for SuperLLM via Mutual Consistency}
\author{Peiwen Yuan$^1$, Shaoxiong Feng$^2$, Yiwei Li$^1$, Xinglin Wang$^1$, Boyuan Pan$^2$\\ {\bf Heda Wang$^2$, Yao Hu$^2$, Kan Li$^{1}$\footnotemark[1]}\\
  $^1$School of Computer Science and Technology, Beijing Institute of Technology \\
  $^2$Xiaohongshu Inc \\
  \texttt{\{peiwenyuan,liyiwei,wangxinglin,likan\}@bit.edu.cn} \\
  \texttt{\{shaoxiongfeng2023,whd.thu\}@gmail.com} \\  \texttt{\{panboyuan,xiahou\}@xiaohongshu.com}}

\begin{document}
\maketitle
\renewcommand{\thefootnote}{\fnsymbol{footnote}} 
\footnotetext[1]{Corresponding author.}

\begin{abstract}
The guidance from capability evaluations has greatly propelled the progress of both human society and Artificial Intelligence. 
However, as LLMs evolve, it becomes challenging to construct evaluation benchmarks for them with accurate labels on hard tasks that approach the boundaries of human capabilities.
To credibly conduct evaluation without accurate labels (denoted as poor-supervised evaluation), we propose the \textsc{PoEM} framework. We first prove that the capability of a model can be equivalently assessed by the consistency between it and certain reference model, when their prediction distributions are independent and the sample size is infinite.
To alleviate the insufficiencies of the conditions in reality, we further introduce an algorithm that treats humans (when available) and the models under evaluation as reference models, alternately conducting model weights calibration and filtering during E-step and M-step.
Comprehensive experiments across 3 types of tasks with 16 mainstream LLMs have shown that \textsc{PoEM} under poor supervision can achieve an average of 0.98 Pearson correlation coefficient with supervised evaluation results, demonstrating good effectiveness, efficiency and generalizability. 
More generally, \textsc{PoEM} has advanced the evaluation paradigm evolution from human-centric to human\&model-centric by treating both of them as reference models, mitigating the limitations of human evaluation in the era of LLMs.
\end{abstract}
\section{Introduction}
\label{sec:intro}


Evaluations, such as IQ tests and Olympic competitions, can effectively identify areas of strength and weakness for individuals or groups, providing valuable guidance for the enhancement of various human abilities. 
In the field of artificial intelligence, benefiting from high-quality annotated data \citep{imagenet,glue}, neural network models have been well evaluated and optimized specifically, leading to significant advancements across various tasks.

Nowadays, large language models (LLMs) \citep{GPT4} have made significant strides in various aspects and shown emergent abilities, even approaching the level of humans in some tasks \citep{supermeaning,boostingreason,summdead,emergentreason}. 
To further guide LLMs towards AGI, accurately evaluating them on a wider range of challenging tasks is of crucial importance.

\begin{figure}
  \vspace*{-10pt}
  \begin{centering}
    \includegraphics[scale=0.34]{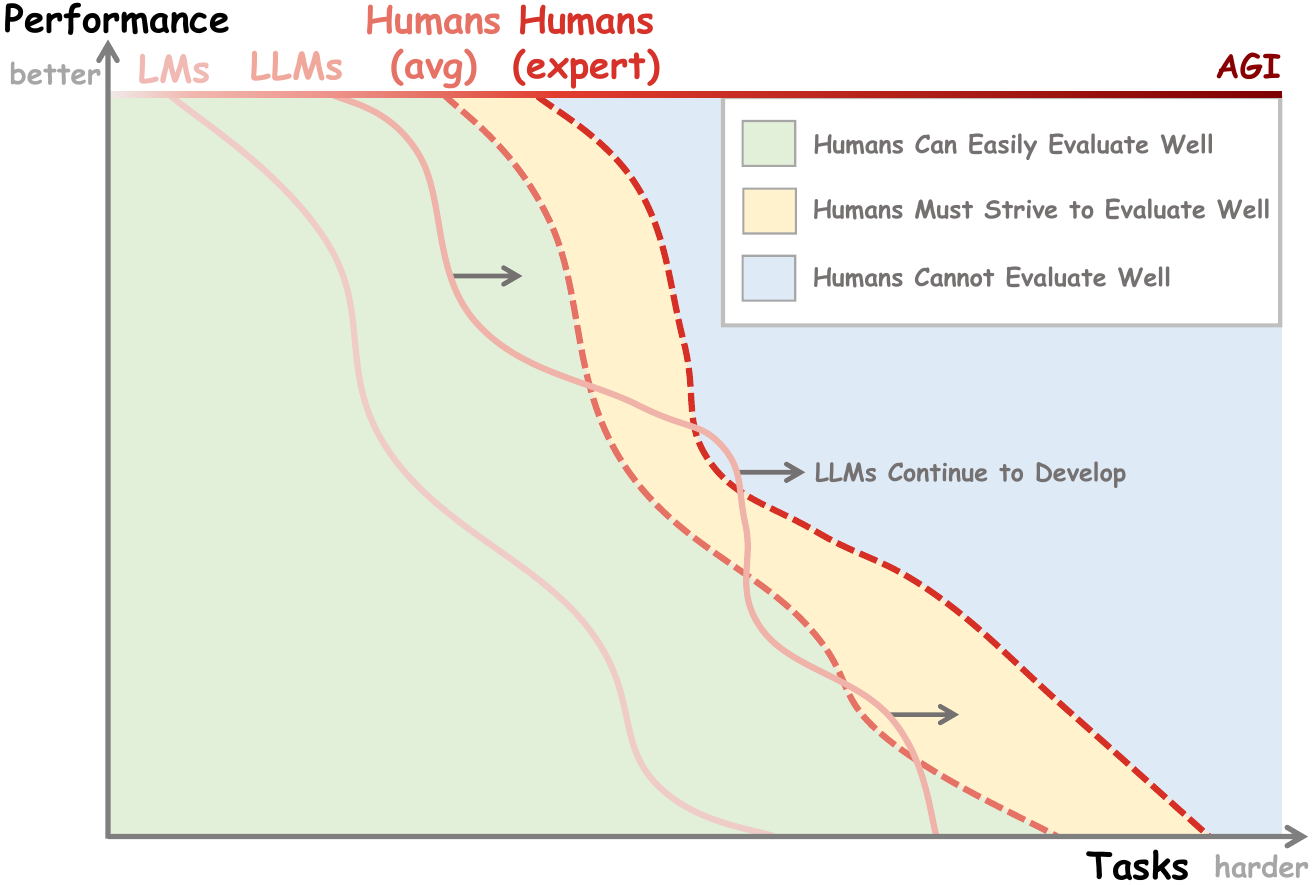}
    \setlength{\abovecaptionskip}{0pt}
    \caption{Schematic diagram of performance of humans and models with varying task difficulty. It is becoming increasingly difficult for humans to accurately evaluate LLMs (offer accurate supervision) with their rapid development.}
    \label{fig:first}
    \setlength{\abovecaptionskip}{0pt}
    \setlength{\belowcaptionskip}{0pt}
  \end{centering}
\end{figure}

However, building benchmarks for challenging tasks is especially intelligence-intensive and sometimes difficult to ensure the accuracy of labels \citep{labelerror,labelerror2}. As shown in Figure~\ref{fig:first}, some tasks (yellow area) require well-trained annotators, or even rare human experts to annotate and verify the labels, which can be difficult to fulfill. 
\citet{usr} show that well-trained annotators can only achieve an average of 0.57 Pearson correlation coefficient when annotating text quality.
More challenging tasks (blue area) may even exceed the capability boundary of experts, such as ``Is the Riemann hypothesis true?'', which cannot be accurately labeled. So here comes the question: \textit{How to evaluate LLMs accurately on benchmarks with poor supervision (with inaccurate labels or even without labels)?}


Some works have explored unsupervised (label-free) model evaluation, including using LLM-based pseudo labels \citep{benchmarking,prd}, evaluating with logits \citep{rel-prob3,rel-prob4}, and detecting conflicts with designed inputs \citep{consevaluate}. Others attempt to correct \citep{correct} or filter \citep{filter} inaccurate labels.
Unfortunately, these approaches present issues such as limited applicability across tasks, logits of some LLMs are unavailable (e.g., GPT-4), and dependence on prior knowledge. 
More importantly, in cases where the true labels for challenging tasks cannot be recovered (beyond the capability boundaries of both humans and models), none of them have theoretically ensured the reliability of the evaluation results.
Fundamentally, there is an urgent need for an LLM evaluation framework that is theoretically grounded, task-agnostic, accurate, and efficient under poor supervision.






Intuitively, examinees can predict their own grades by examining their mutual consistencies. Inspired by this, we propose the \textsc{PoEM} framework (\textbf{Po}or-supervised \textbf{E}valuation with \textbf{M}utual Consistency). We first derive the theoretical conditions needed to represent model capability with inter-model consistencies. Afterwards, we introduce an algorithm to alleviate the impact of the insufficient conditions in real-world scenarios, thereby facilitating credible evaluations under poor supervision.

Specifically, we derive that when the sample size is infinite, if a reference model performs better than random guessing and its predictions are independent of the models being evaluated, then the consistency with this reference model can accurately reflect model capability.
Preliminary experiments demonstrate that randomly picking a model as the reference fails to meet these conditions in real-world scenarios. To address such issues, we introduce an algorithm through a step-by-step construction process. It regards all the models under evaluation and humans (who may provide inaccurate labels, optional for the algorithm) as reference models to mitigate the limitation of sample size. On this basis, It conducts calibration and filtering operations alternately during E-step and M-step to alleviate the potential dependency among model predictions.

\textsc{PoEM} framework holds several advantages.
\textbf{Theoretically Grounded}: Theorem~\ref{theorem1} forms the foundation of \textsc{PoEM}.
\textbf{Task-Agnostic}: It can be applied across various tasks, as we demonstrate in \S\ref{exp}.
\textbf{Accurate}: Experimental results show that \textsc{PoEM} under poor supervision aligns perfectly with supervised evaluation results, achieving an average of 0.98 Pearson and 0.97 Spearman correlation coefficients.
\textbf{Efficient}: No additional inference overhead is required except for model predictions.

Furthermore, when human annotations are available, we validate that \textsc{PoEM} allows for a more accurate evaluation results by treating both of humans and models as reference models, as opposed to the traditional consistency check between model predictions and human annotations.
From this perspective, \textsc{PoEM} has advanced the evaluation paradigm evolution from human-centric to human\&model-centric, mitigating the limitations of human evaluation capabilities in the era of LLMs.

Our contributions are summarized as follows:

\begin{itemize}[leftmargin=20pt]
\setlength{\itemsep}{0pt}
\setlength{\parsep}{0pt}
\setlength{\parskip}{0pt}
\item We discuss the significance of evaluating LLMs under poor supervision with their continuing development and propose the \textsc{PoEM} framework for this scenario.
\item We theoretically derive that mutual consistency of models can accurately assess model capabilities under certain conditions.
\item We propose an algorithm to significantly alleviate the insufficiencies of the conditions in real-world scenarios for credible poor-supervised evaluation.
\item We experimentally validate the good effectiveness, efficiency and generalizability of \textsc{PoEM} across regression, classification, and reasoning tasks with 16 mainstream LLMs.
\end{itemize}

\section{Related Work}
Surrounding our study, we discuss researches on poor-supervised evaluation and model consistency.
\paragraph{Poor-supervised Evaluation}
We refer to the process of evaluating models on benchmarks that contain inaccurately labeled or unlabeled data as poor-supervised evaluation, an area where many related efforts have been invested.
Various directions have been explored to assessing model's capabilities on datasets without labels: examining distribution discrepancy \citep{rel-prob3, rel-dis1} , relying on model confidence \citep{rel-confi1,rel-confi2,rel-confi3,rel-prob4}, calculating models' disagreements \citep{rel-iid,rel-dis2,rel-sameori2,rel-sameori1} and bucketing based on decision boundaries \citep{rel-buck1,rel-buck2,rel-buck3}. 
Studies of another direction \citep{rel_cor1,rel_cor2,rel_cor3,rel_cor4} consider filtering or correcting inaccurate labels.
However, these studies mainly focus on classification tasks and often require extra model training or model logits, thus lacking generalizability and not being applicable to LLMs whose logits are unavailable.
In addition, \citet{consevaluate} and \citet{bringdata} consider detecting conflicts with designed inputs to assess model's capabilities. However, the design of input format highly depends on prior knowledge and cannot be generalized across tasks.
Conducting self-evaluation \citep{judging} or peer evaluation \citep{prd,benchmarking} with LLMs is a promising direction, but current studies have not addressed the fundamental issue of how to ensure the reliability of LLM-evaluators on benchmarks that exceed their own capabilities. 
Additionally, such methods require extra inference overhead.
Overall, a theoretically grounded, accurate, task-agnostic and efficient poor-supervised evaluation framework for LLMs remains to be researched, for which we propose \textsc{PoEM}.
\paragraph{Model Consistency}
Consistency has long been studied for training models \citep{rel-cons1,rel-cons2}, enhancing performance during inference \citep{SC,rel-cons5,SC1,SC2,SC3}, and examining specific attributes such as reliability \citep{rel-cons4} and hallucination \citep{rel-cons8}. We investigate using mutual consistency to conduct poor-supervised evaluation.





\section{\textsc{PoEM} Framework}
In this section, we first introduce the task definition of poor-supervised evaluation in \S\ref{sec:task_def}. We then present the foundational Theorem~\ref{theorem1} of the \textsc{PoEM} framework in \S\ref{sec:the}. Afterwards, we analyze the applicability of naively applying Theorem~\ref{theorem1} in real-world scenarios through preliminary experiments in \S\ref{sec:pre}. Finally, in \S\ref{sec:met}, we propose the algorithm to maximize the satisfaction of the conditions required by Theorem~\ref{theorem1}, thus achieving credible poor-supervised evaluation.

\subsection{Task Definition}
\label{sec:task_def}
Given LLMs $\{\mathcal{M}^i\}^L_{i=1}$, datasets $\mathbb{D}: (X=\{x_i\}^N_{i=1},Y=\{y_i\}^N_{i=1})$ and predictions of $\mathcal{M}^i$ on $X$ as $\hat{Y}^i=\{\hat{y}^i_j\}^N_{j=1}$, the consistency between $\{\hat{Y}^i\}_{i=1}^L$ and true labels $Y$, denoted as $\{Cons(\hat{Y}^i,Y)\}^L_{i=1}$, shortened to $B \in \mathcal{R}^{L}$, is the accurate measurement of the capabilities of $\{\mathcal{M}^i\}^L_{i=1}$.
Poor-supervised evaluation requires an algorithm $\mathcal{F}$ that can output $\hat{B}\in \mathcal{R}^{L}$ when $Y$ is unavailable, where $\hat{B}_i$ can substitute for $B_i$ in evaluating the capability of $\mathcal{M}^i$. The correlation coefficient between $\hat{B}$ and $B$ can be calculated to validate the effectiveness of $\mathcal{F}$.

\subsection{Equating Mutual Consistency with Capability}
\label{sec:the}
\newtheorem{theorem}{Theorem}
\begin{theorem}
\label{theorem1}
When the size of $X$ is infinite (Condition 1), if certain reference model $\dot{\mathcal{M}}$ performs better than random guessing (Condition 2) and its predictions are independent of models $\{\mathcal{M}^i\}_{i=1}^L$ under evaluation (Condition 3), the following equation holds (See Appendix~\S\ref{theorem_proof} for derivation):
\begin{equation}
\setlength{\jot}{2pt}
\small
\begin{gathered}
Cons(\mathcal{M}^i(X),\dot{\mathcal{M}}(X))<Cons(\mathcal{M}^j(X),\dot{\mathcal{M}}(X))
\Leftrightarrow B_i<B_j
\end{gathered}
\label{equation: div-re-detail}
\end{equation}
where $Cons(\cdot)$ denotes mutual consistency (See calculation methods in \S\ref{exp} for different tasks).
\end{theorem}
Theorem~\ref{theorem1} allows us to credibly evaluate and compare the model capabilities without true labels $Y$. The intuition behind Theorem~\ref{theorem1} is straightforward to understand: For the samples where the reference model $\dot{\mathcal{M}}$ predicts correctly, strong model tend to have higher consistency with the reference model compared to weak model, $Cons(\mathcal{M}^{weak},\dot{\mathcal{M}})$$<Cons(\mathcal{M}^{strong},\dot{\mathcal{M}})$. While for the samples where the reference model predicts incorrectly, if the reference model does not tend to make the same predictions as either strong or weak models, the consistency of both strong and weak model with the reference model should be the same, $Cons(\mathcal{M}^{strong},\dot{\mathcal{M}}) \approx Cons(\mathcal{M}^{weak},\dot{\mathcal{M}})$. Therefore, across all the samples, $Cons(\mathcal{M}^{weak},\dot{\mathcal{M}})$$<Cons(\mathcal{M}^{strong},\dot{\mathcal{M}})$.

\begin{figure*}[!htb]
    \centering
    \subfigure[Consistency Matrix]{\includegraphics[width=0.49\hsize, height=0.38\hsize]{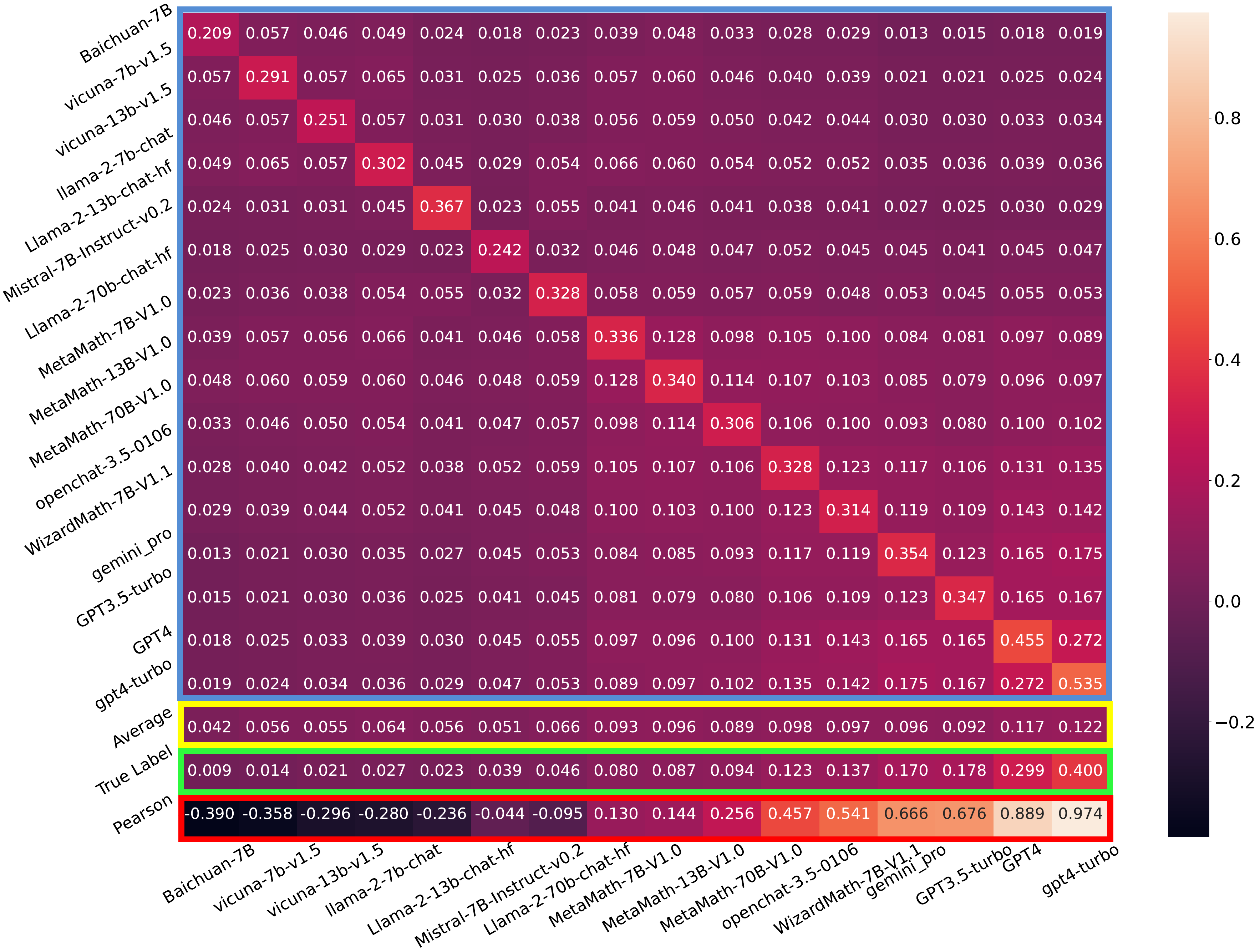}\label{fig: sub_figure1}} 
    \subfigure[Affinity Matrix]{\includegraphics[width=0.49\hsize, height=0.38\hsize]{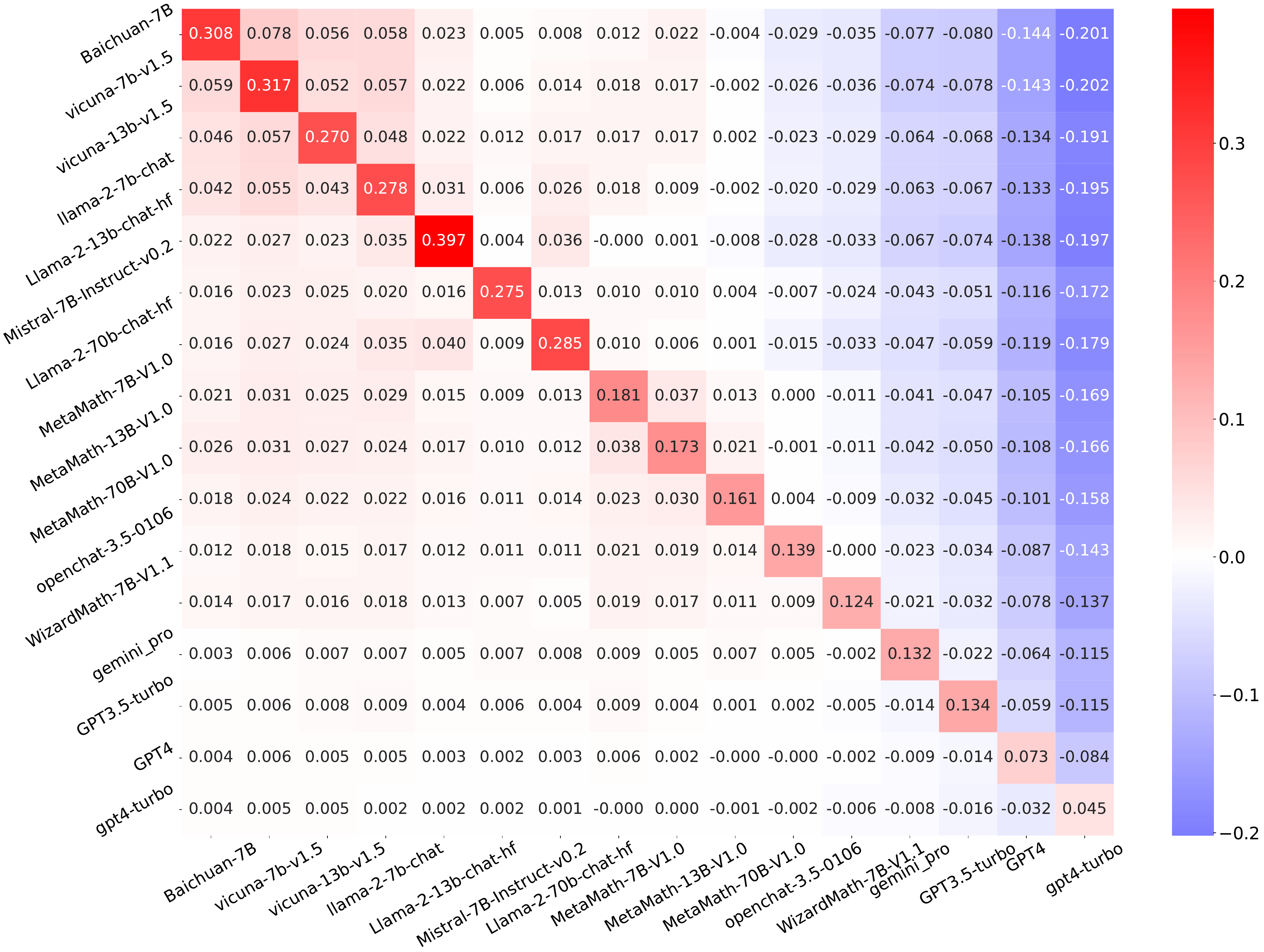} \label{fig: sub_figure2}}
    \setlength{\belowcaptionskip}{0pt}
    \caption{Mutual consistency and affinity matrices on MATH-Precalculus dataset among LLMs.}
    \label{fig:explore}
\end{figure*}

\subsection{Results and Insights of Preliminary Experiments}
\label{sec:pre}
In real scenarios, LLMs under evaluation can serve as reference models. However, although Condition 2 can naturally be assumed to be satisfied, Condition 1 and 3 cannot be completely fulfilled since the size of $X$ is finite and some LLMs may exhibit mutual affinity and tend to make the same predictions (e.g., Llama series). To explore the impact of the insufficiencies of Condition 1 and 3 in real scenarios, we visualize the consistency matrix $C$ and the affinity matrix $A$ of models to be evaluated.
$C_{i,j}$ $\in$ [0,1] denotes $Cons(\hat{Y}^i,\hat{Y}^j)$,
and $A_{i,j}$ $\in$ [-1,1] reflects the affinity of $\mathcal{M}^i$ to $\mathcal{M}^j$:
\begin{equation}
\label{eq:pre-1}
\small
A_{i,j} = \frac{C_{i,j}}{Sum(C_{i})} - \frac{B_j}{Sum(B)}
\end{equation}
A positive $A_{i,j}$ implies picking $\mathcal{M}^i$ as $\dot{\mathcal{M}}$ will overestimate the capability of $\mathcal{M}^j$ and vice versa.

We conduct this preliminary experiment on subset Precalculus of MATH \citep{MATH} benchmark with 16 LLMs (See detailed setup in \S\ref{sec:exp1setup}). 
As shown in Figure~\ref{fig: sub_figure1}, the blue rows demonstrate the $C$, the yellow row calculates $\{Avg(C_i)\}_{i=1}^L$, the green row shows $B$, and the $i^{th}$ column of the red row represents the Pearson coefficient between $C_i$ and $B$ (the closer to 1, the stronger the linear correlation). 
From Figure~\ref{fig: sub_figure1}, we can indicate from the red row that randomly picking an LLM as the reference model can lead to highly unstable and potentially negatively correlated evaluation outcomes due to the insufficiencies of conditions. Further, we can observe the following tendencies and corresponding insights.
(1) The yellow and green rows show a positive correlation $\rightarrow$ \textit{\textbf{Insight 1}}: \textit{stronger models tend to have higher consistency with other models};
(2) The red and green rows also show a positive correlation $\rightarrow$ \textit{\textbf{Insight 2}}: \textit{stronger models' consistency with other models can better reflect the true performance of these models.}
Considering that model predictions tend to converge towards the true labels and predictions of stronger models are closer to the true labels, the above two insights are natural and reasonable.

From Figure~\ref{fig: sub_figure2}, we confirm that some LLMs belonging to the same series (Llama-2 series, MetaMath series) indeed exhibit mutual affinity, which makes their consistency biased to reflect their true performance. Furthermore, we observe \textit{\textbf{Insight 3}}: \textit{weaker models tend to exhibit higher affinity.} We attribute this to the consistency between weak models and strong models underestimating the performance of the strong models, indirectly leading to a higher affinity among weak models. More generally, the proportion of samples correctly predicted by a weak model is too low, using it as a reference model may not suffice to differentiate between strong and weak models.


\subsection{Algorithms for Mitigating Insufficient Conditions}
\label{sec:met}
\begin{figure*}[t]
\centering
\includegraphics[width=0.96\textwidth]{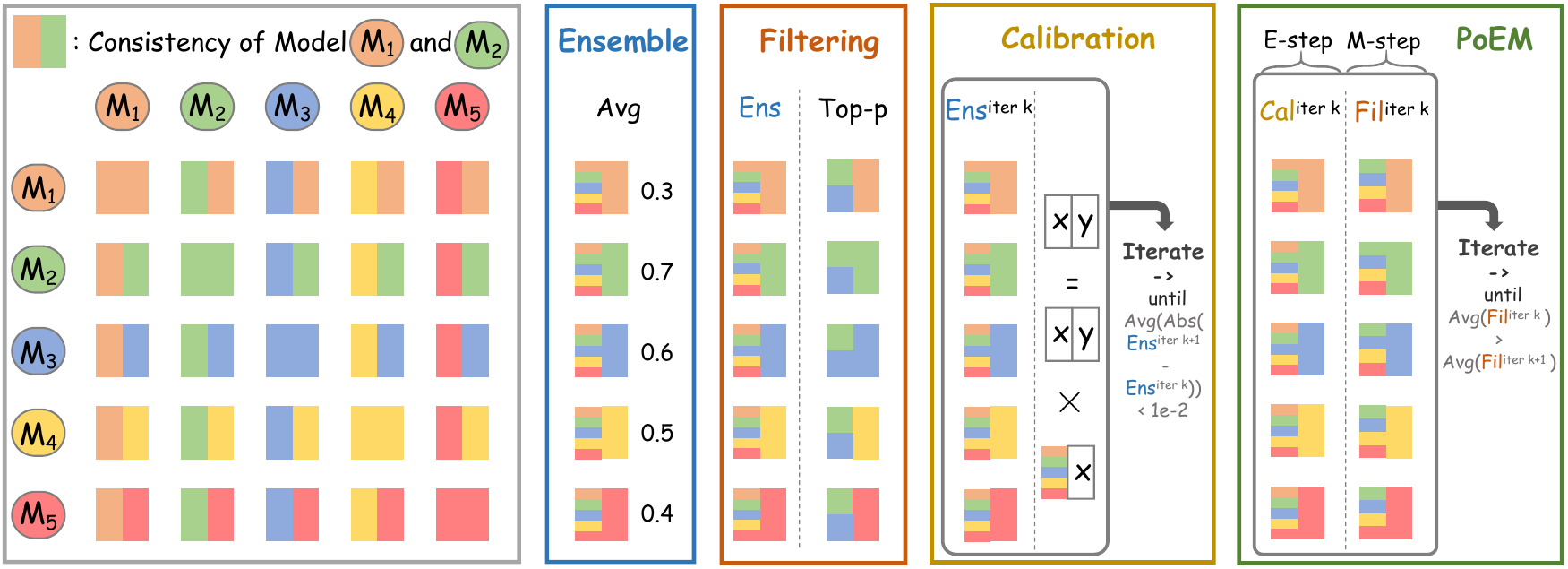}
\caption{Overall illustration of our proposed algorithms.}
\label{fig:main}
\end{figure*}

Since randomly picking an LLM as the reference model can lead to unstable evaluation results, we introduce the algorithm $\mathcal{F}_{\textsc{PoEM}}: C \rightarrow \hat{B}$ through a step-by-step construction process to better meet the conditions of Theorem~\ref{theorem1} for more accurate $\hat{B}$ based on the insights above, as shown in Figure~\ref{fig:main}. 

\subsubsection{Naive Ensemble}
Viewing each (reference model, sample) couple as an evaluation data point, we can easily enlarge sample size by $L$ times by regarding all the LLMs as reference models:
\begin{equation}
\label{eq:met-1}
\small
\mathcal{F}_{ensemble}(C) = \{Avg(C_{i})\}_{i=1}^L
\end{equation}
where $Avg(C_{i})$ denotes the ensemble consistencies of certain model $\mathcal{M}^i$ with all the models.
Compared with randomly picking certain model as the reference model, $\mathcal{F}_{ensemble}$ can offer a more stable measurement result by alleviate the insufficiency of sample size (Condition 3).



\subsubsection{Weight Calibration}
Building upon $\mathcal{F}_{ensemble}$, considering that $C_i$ of stronger model $\mathcal{M}^i$ aligns better with $B$ (\textit{\textbf{Insight 2}}), we contemplate calibrating the ensemble weights based on models capabilities through iteration. 

Let $\alpha_j^k$ be the weight assigned to $C_{i,j}$ after iteration $k$, which is initialized as $1/L$. The capability estimation of $\mathcal{M}^i$ for iteration $k$, $\hat{B}^k_i$, is the calibrated average of its consistencies with all the models:
\begin{equation}
\label{eq:met-2}
\small
\hat{B}^k_i = \sum_{j=1}^L C_{i,j} \times \alpha_j^{k-1}
\end{equation}
For each iteration $k$, we calibrate $\alpha_j^k$ according to $\hat{B}^k$ following \textit{\textbf{Insight 2}}:
\begin{equation}
\label{eq:met-3}
\small
\alpha_j^k = \frac{\hat{B}^k_j}{\sum_{i=1}^L \hat{B}^k_i}
\end{equation}
Given the set of equations above, we look for the converging point after $\hat{k}$ iterations where the following equation holds:
\begin{equation}
\label{eq:met-4}
\small
\frac{1}{L}\sum_{j=1}^L |\alpha_j^{\hat{k}+1}-\alpha_j^{\hat{k}}|<1e-2
\end{equation}
Through the above process, we have managed to calibrate the ensemble weights using the capability estimation of the models and obtaining the outcomes of algorithm $\mathcal{F}_{calibrate}$:
\begin{equation}
\label{eq:met-5}
\small
\mathcal{F}_{calibrate}(C) = \{\sum_{j=1}^L C_{i,j} \times \alpha_j^{\hat{k}}\}_{i=1}^L
\end{equation}

\subsubsection{Reference Model Filtering}
To alleviate the impact of non-independent distributions between model predictions (insufficiency of Condition 1) shown in Figure~\ref{fig: sub_figure2}, we consider filtering out models with strong affinity tendencies. \textit{\textbf{Insight 3}} indicates that weak models generally have high affinity tendencies, thus we contemplate filtering them from reference models based on the estimation of their capabilities according to $\mathcal{F}_{ensemble}$ as follows:
\begin{gather}
\label{eq:met-6}
\hat{B}^{ens} = \mathcal{F}_{ensemble}(C) \\
\mathcal{S}_{ref} = \mathrm{arg}\{i|  \hat{B}^{ens}_i > Max(\hat{B}^{ens})\times p\} \\
\mathcal{F}_{filter} (C) = \{Avg(\{C_{i,j}\}_{j \in {\mathcal{S}_{ref}}})\}_{i=1}^L
\end{gather}
The hyperparameter $p$ serves as a threshold for filtering out weak models whose performance is not up to par with the strongest model to a certain extent according to $\hat{B}^{ens}$. Through the above process, the weak model with high affinity tendencies can be filtered for calculating a debiased $\hat{B}$.

\subsubsection{EM-based Integration}
Based on the discussions above, we consider designing an algorithm that can integrate all three insights to mitigate the insufficiencies of $Condition\ 1$ and $3$. To this end, we propose an EM-based \citep{em} algorithm $\mathcal{F}_{\textsc{PoEM}}$ as follows:
\begin{equation}
\label{eq:met-7}
\small
\mathcal{F}_{\textsc{PEEM}}(C) = \{\sum_{j=1}^L C_{i,j} \times \alpha_j^{\hat{k}}\times \beta_j^{\hat{k}}\}_{i=1}^L
\end{equation}
We set $\beta_j \in \{0,1\}$ to determine whether $\mathcal{M}^j$ serves as a reference model and $\alpha_j \in [0,1]$ to control the weight $\mathcal{M}^j$ serving as the reference model.
\paragraph{Initialization.} We initialize $\alpha_i^0$ as $\frac{1}{L}$ and $\beta_i^0$ as 1.
\paragraph{Optimizing Objective.} 
From \textit{\textbf{Insight 1}} and \textit{\textbf{2}}, we deduce the following chain-of-thought: greater greater $Avg(C_i) \xleftrightarrow{\text{Insight 1}}$ stronger $M_i \xleftrightarrow{\text{Insight 2}}$ better alignment between $C_i$ and $B$. Inspired by this, $Avg(\hat{B})$ can serve as a proxy optimizing objective for $\hat{B}$ to better approximate $B$.

\paragraph{Expectation Step.} In this step, we conduct $\mathcal{F}_{calibrate}$ among the reference model set constructed by $\beta^{k-1}$ to obtain the calibrated expectation of $\alpha^k$ as follows:
\begin{gather}
\label{eq:met-8}
\mathcal{S}_{ref}^{k-1} = \mathrm{arg}\{ i| \beta^{k-1}_i = 1\} \\
\{\alpha^k_i\}_{i\in \mathcal{S}_{ref}^{k-1}} = \mathcal{F}_{calibrate}(C_{i\in \mathcal{S}_{ref}^{k-1},j\in \mathcal{S}_{ref}^{k-1}}) 
\end{gather}
\paragraph{Maximization Step.} Further, we optimize $\beta^{k}$ to maximize our objective $Avg(\hat{B}^k)$ as follows:
\begin{gather}
\label{eq:met-9}
\small
Index_{min}^k = \underset{i}{\mathrm{argmin}}\ \alpha^k_i \\
\beta^k_i = \begin{cases}
\beta^{k-1}, \ i \neq Index_{min}^k
 \\
0, \ i = Index_{min}^k
\end{cases}\\
Avg(\hat{B}^k) = \frac{1}{\sum_{i=1}^L \beta_i^k} \sum_{i=1}^L \sum_{j=1}^L C_{i,j} \times \alpha_j^{k}\times \beta_j^{k}
\end{gather}
\paragraph{Termination condition.} We iteratively conduct the E-step and M-step for $\hat{k}+1$ rounds until the following formula holds true for the first time:
\begin{equation}
\label{eq:met-10}
\small
Avg(\hat{B}^{\hat{k}})>Avg(\hat{B}^{\hat{k}+1})
\end{equation}

\begin{table*}[hb]
    \caption{Model-level Pearson ($r_p$) / Spearman ($r_s$) correlations of different algorithms on 7 subsets of MATH. All results are statistically significant (p-value < 0.05).}\label{tab:exp1}
    \begin{center}
    \begin{small}
    \begin{sc}
    \setlength{\tabcolsep}{0.13em} 
    \begin{tabular}{lccp{0.005pt}ccp{0.005pt}ccp{0.005pt}ccp{0.005pt}ccp{0.005pt}ccp{0.005pt}ccp{0.005pt}cc}
    \toprule
    \multirow{2}{*}{\textbf{Subset}} &\multicolumn{2}{c}{\textbf{IA}}&& \multicolumn{2}{c}{\textbf{PA}} && \multicolumn{2}{c}{\textbf{GE}} && \multicolumn{2}{c}{\textbf{CP}} && \multicolumn{2}{c}{\textbf{AG}}&& \multicolumn{2}{c}{\textbf{NT}}&& \multicolumn{2}{c}{\textbf{PC}}&& \multicolumn{2}{c}{\textbf{Average}} \\
    & $\bm{r_p}$\ &\ $\bm{r_s}$&&$\bm{r_p}$\ &\ $\bm{r_s}$&&$\bm{r_p}$\ &\ $\bm{r_s}$&&$\bm{r_p}$\ &\ $\bm{r_s}$&&$\bm{r_p}$\ &\ $\bm{r_s}$&& $\bm{r_p}$\ &\ $\bm{r_s}$&& $\bm{r_p}$\ &\ $\bm{r_s}$&& $\bm{r_p}$\ &\ $\bm{r_s}$ \\   
    \midrule
    \textcolor{gray}{Random Pick}&\textcolor{gray}{.145}&\textcolor{gray}{.336}&&\textcolor{gray}{.439}&\textcolor{gray}{.646}&&\textcolor{gray}{.264}&\textcolor{gray}{.495}&&\textcolor{gray}{.301}&\textcolor{gray}{.511}&&\textcolor{gray}{.410}&\textcolor{gray}{.620}&&\textcolor{gray}{.259}&\textcolor{gray}{.515}&&\textcolor{gray}{.125}&\textcolor{gray}{.325}&&\textcolor{gray}{.278}&\textcolor{gray}{.492}\\
    Ensemble&.739&.791&&.964&.992&&.915&.970&&.873&.983&&.952&.994&&.859&.969&&.793&.866&&.871&.938\\
    Calibration&.784&.915&&.941&.992&&.883&.984&&.850&.992&&.921&.993&&.846&.985&&.826&.969&&.864&.976\\
    Filtering&.834&.886&&\textbf{.997}&\textbf{.998}&&\textbf{.986}&.990&&.980&.994&&.995&.999&&.969&.983&&.889&.931&&.950&.969\\
    \textsc{PEEM}&\textbf{.912}&\textbf{.958}&&.994&.997&&.978&.992&&\textbf{.985}&\textbf{.996}&&\textbf{.996}&\textbf{1.00}&&\textbf{.977}&\textbf{.991}&&\textbf{.955}&\textbf{.987}&&\textbf{.971}&\textbf{.989}\\
    \bottomrule
    \end{tabular}
    \end{sc}
    \end{small}
    \end{center}
\end{table*}

In summary, we alternately calibrate the model weights and filter out weak models with high affinity tendency according to the estimation under current parameter, iterating this process until the optimal proxy objective $Avg(\hat{B})$ is achieved.


\section{Experiments}
\label{exp}

In this section, we conduct experiments using 16 mainstream LLMs (Table~\ref{tab:exp0}) with a broad range of size and capabilities across 3 distinct task categories to comprehensively validate the \textsc{PoEM} framework.
To cover different scenarios under poor-supervised evaluation, we explore settings without labels in reasoning (\S\ref{exp1}) and regression (\S\ref{exp2}) tasks, and settings with inaccurate labels in classification (\S\ref{exp3}) task.

\subsection{Reasoning Task}
\label{exp1}
We first conduct experiments on MATH \citep{MATH} benchmark, which contains 7 subsets (Table~\ref{tab:exp2}), to assess the reasoning capabilities of LLMs without labels.

\subsubsection{Experimental Setup}
\label{sec:exp1setup}
To enhance the stability of evaluation, for model $\mathcal{M}^i$ and sample $x_j$, we sample $T(\hat{Y}^i)$ ($=5$ across experiments) times at temperature 0.5 to attain predictions $\{\hat{y}_{j,k}^i\}_{k=1}^{T(\hat{Y}^i)}$ (See \S\ref{sec:greedy} for results of greedy search). As the domain of true label $y$ is discrete in MATH, we calculate $Cons(Y,\hat{Y}^i)$ as follows:
\begin{equation}
    \small
    \label{ex:eq1}
    Cons(Y,\hat{Y}^i) = \frac{1}{NT(\hat{Y}^i)T(Y)}\sum_{j=1}^N\sum_{u=1}^{T(Y)}\sum_{v=1}^{T(\hat{Y}^i)}\mathbf{1}_{y_{j,u}=\hat{y}_{j,v}^i}
\end{equation}
The mutual consistency $Cons(\hat{Y}^i,\hat{Y}^j)$ can be similarly calculated. $T(Y)=1$ as the label is unique for each $x_i$. To enhance the significance of the results, we randomly sample models at a ratio of $ q_{model}=0.7 $ from 16 candidates and take the mean of 500 outcomes to report. As for $\mathcal{F}_{filter}$, we set filtering threshold $p$ as 0.9 in default (See Appendix~\S\ref{sec:p} for results with varying $p$). 
Apart from the proposed algorithms, we include randomly picking a reference model from LLMs for baseline comparison.
The Pearson and Spearman coefficient, $r_p$ and $r_s$, can separately measure the degree of linear correlation and the monotonic relationship between two variables. We use them to measure the correlation between $\hat{B}$ and $B$, which are separately obtained from our algorithm and the true labels.


\subsubsection{Experimental Results}
As shown in Table~\ref{tab:exp1}, \textsc{PoEM} achieves correlation coefficient greater than 0.9 across all the subsets, indicating that it can assess model capabilities well through mutual consistency without true labels. 

\paragraph{Horizontal Comparison across Algorithms.} $\mathcal{F}_{ensemble}$ performs much better than randomly picking an LLM to be the reference model as it can fully use the predictions from all the LLMs, indirectly alleviating the limitation of sample size (Condition 3). 
$\mathcal{F}_{calibrate}$ further improves the ordering of model capability (higher $r_s$) by calibrating the model weights during ensemble. 
However, altering the original weights also leads to a slight decrease in the linear relationship (lower $r_p$). By filtering out weak models with high affinity tendency on the basis of $\mathcal{F}_{ensemble}$ to mitigate Condition 1, $\mathcal{F}_{filter}$ achieves significant improvements on both $r_s$ and $r_p$. 
Finally, by integrating the advantages of the above methods through EM algorithm, $\mathcal{F}_{PoEM}$ significantly mitigates the insufficient conditions of Theorem~\ref{theorem1}, achieving evaluation results that align perfectly with supervised evaluation (0.971 $r_p$ and 0.989 $r_s$). 

\begin{figure}
  \begin{centering}
    \includegraphics[scale=0.36]{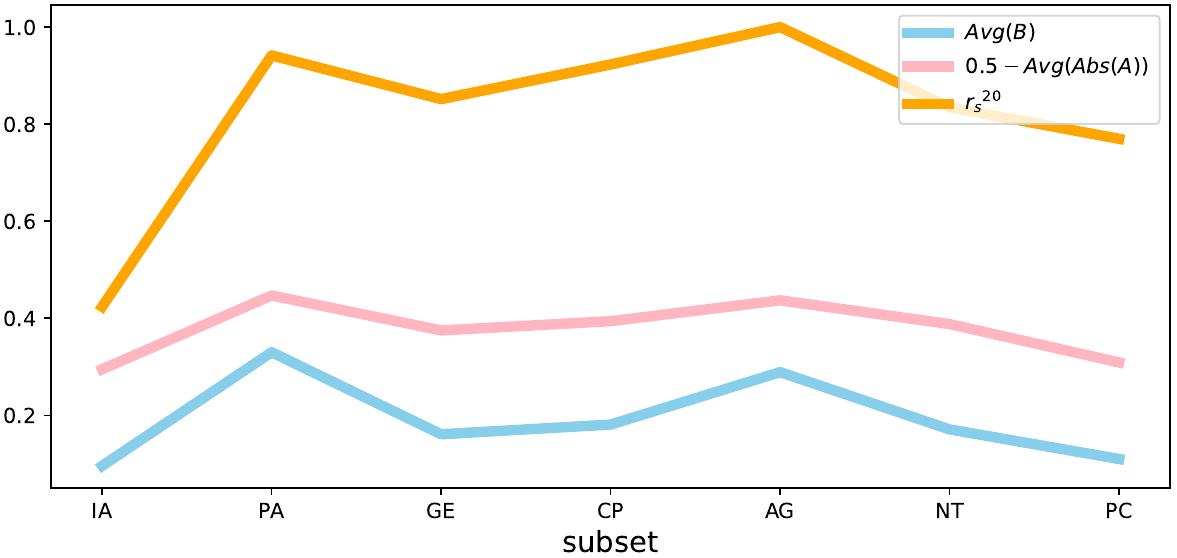}
    \setlength{\abovecaptionskip}{0pt}
    \caption{Comparisons between $r_s$ of \textsc{PoEM}, the mean of the absolute values of affinity matrix $Avg(Abs(A))$, and average accuracy of all the models $Avg(B)$. We plot the $20^{th}$ power of $r_s$ and $0.5-Avg(Abs(A))$ for easier observation.}\label{fig:acc_pea}
    \setlength{\abovecaptionskip}{0pt}
    \setlength{\belowcaptionskip}{0pt}
  \end{centering}
\end{figure}


\paragraph{Analyses of Varying Results across Subsets.} 
As we notice that the performance of the proposed algorithms vary across subsets, we analyze the underlying reason by comparing the average accuracy of all the models $Avg(B)$, the mean of the absolute values of affinity matrix $Avg(Abs(A))$, and $r_s$ of the last row in Table~\ref{tab:exp1}. We plot the $20^{th}$ power of $r_s$ and $0.5-Avg(Abs(A))$ for easier observation. 
As shown in Figure~\ref{fig:acc_pea}, they exhibit a clear correlation. We speculate the reasoning chain is as follows: on subsets where the model accuracy ($Avg(B)$) is higher, their predictions tend to be consistent with true labels, showing a weak affinity tendency ($Avg(Abs(A))$). Therefore, $Condition\ 2$ can be better satisfied for $\mathcal{F}_{PoEM}$ to attain good performance ($r_s$), and vice versa. 
Inspired by this, we suggest that, where feasible, samples of moderate difficulty rather than those obviously beyond the model's capability should be chosen to evaluate the model, thereby making the evaluation results more accurate. 


\subsection{Regression Task}
\label{exp2}
We select the USR benchmark \citep{usr} within the context of regression task. 

\begin{table*}[hb]
    \vskip 0.15in
    \caption{Model-level Pearson ($r_p$) / Spearman ($r_s$) correlations of different algorithms with different sample sampling ratios $q_{sample}$ on USR benchmark. P-values of all the results < 0.05.}\label{tab:exp3}
    \begin{center}
    \begin{small}
    \begin{sc}
    \setlength{\tabcolsep}{0.20em} 
    \begin{tabular}{lccp{0.005pt}ccp{0.005pt}ccp{0.005pt}ccp{0.005pt}ccp{0.005pt}cc}
    \toprule
    \multirow{2}{*}{$\bm{q_{sample}}$} &\multicolumn{2}{c}{\textbf{0.1}}&&\multicolumn{2}{c}{\textbf{0.2}}&& \multicolumn{2}{c}{\textbf{0.4}} && \multicolumn{2}{c}{\textbf{0.6}} && \multicolumn{2}{c}{\textbf{0.8}} && \multicolumn{2}{c}{\textbf{1}} \\
    & $\bm{r_p}$\ &\ $\bm{r_s}$&&$\bm{r_p}$\ &\ $\bm{r_s}$&&$\bm{r_p}$\ &\ $\bm{r_s}$&&$\bm{r_p}$\ &\ $\bm{r_s}$&&$\bm{r_p}$\ &\ $\bm{r_s}$&& $\bm{r_p}$\ &\ $\bm{r_s}$ \\   
    \midrule
    Ensemble&.824&.788&&.867&.836&&.890&.851&&.901&.854&&.902&.859&&.909&.868\\
    Calibration&.819&.805&&.869&.894&&.871&.984&&.903&.883&&.908&.885&&.913&.883\\
    Filtering&.852&.830&&.898&.885&&.920&.908&&.929&.912&&.933&.917&&.937&.920\\
    \textsc{PEEM}&\textbf{.878}&\textbf{.852}&&\textbf{.921}&\textbf{.895}&&\textbf{.952}&\textbf{.928}&&\textbf{.957}&\textbf{.937}&&\textbf{.961}&\textbf{.942}&&\textbf{.966}&\textbf{.942}\\
    \bottomrule
    \end{tabular}
    \end{sc}
    \end{small}
    \end{center}
\end{table*}

\begin{figure}
  \begin{centering}
    \includegraphics[scale=0.36]{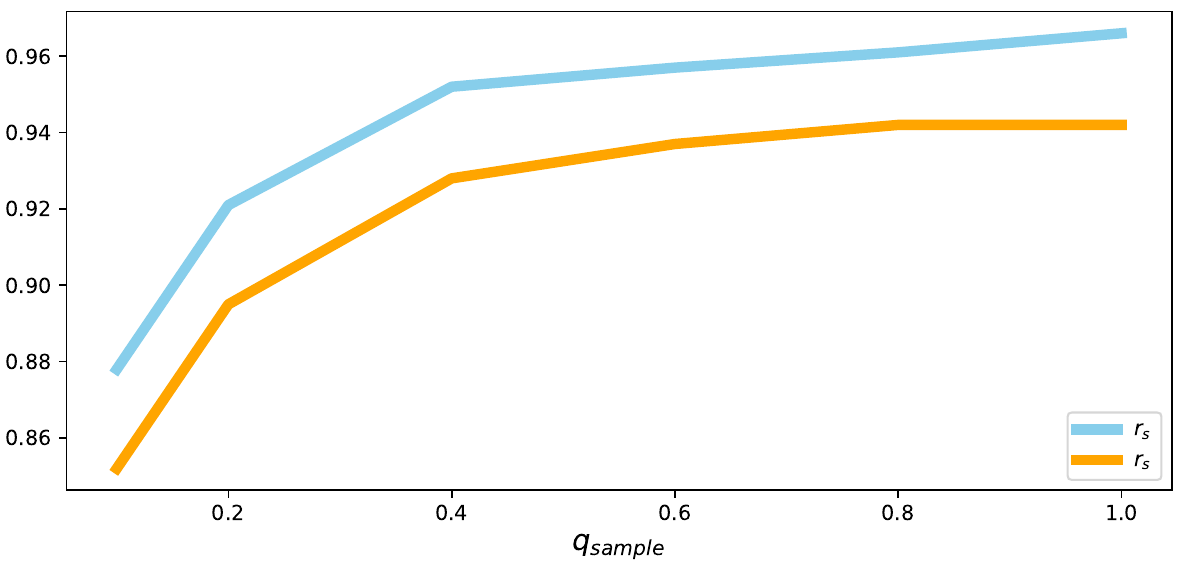}
    \setlength{\abovecaptionskip}{0pt}
    \caption{$r_p$ and $r_s$ with error bar of \textsc{PoEM} as sample size changes.}\label{fig:acc_pea_sample}
    \setlength{\abovecaptionskip}{0pt}
    \setlength{\belowcaptionskip}{0pt}
  \end{centering}
\end{figure}

\subsubsection{Experimental Setup}
USR is a dialog evaluation testbed requiring models to predict quality scores based on specific criteria for the provided dialogues. We follow \citet{geval} to calculate $Cons(Y,\hat{Y}^i)$ as shown below:
\begin{equation}
    \small
    \label{ex:eq2}
    Cons(Y,\hat{Y}^i) = r_p(Y,\hat{Y}^i)
\end{equation}
We choose the \texttt{Overall} criteria for experiments, which contains 360 samples. To validate the effectiveness of the proposed methods with varying sample sizes, we conduct experiments by sampling samples at ratios $q_{sample} \in [0.1,0.2,0.4,0.6,0.8,1]$.

\subsubsection{Experimental Results}
As shown in Table~\ref{tab:exp3}, \textsc{PoEM}  achieves 0.966 $r_p$ and 0.942 $r_s$ when $q_{sample}=1$, demonstrating good effectiveness and generalizability on regression task. 

\begin{table*}[t]
    \caption{Model-level Pearson ($r_p$) / Spearman ($r_s$) correlations of different algorithms with different sample sampling ratios $q_{sample}$ on MultiRC benchmark. P-values of all the results < 0.05.}
    \label{tab:exp4}
    \begin{center}
    \begin{small}
    \begin{sc}
    \setlength{\tabcolsep}{0.20em} 
    \begin{tabular}{lccp{0.005pt}ccp{0.005pt}ccp{0.005pt}ccp{0.005pt}ccp{0.005pt}cc}
    \toprule
    \multirow{2}{*}{$\bm{q_{sample}}$} &\multicolumn{2}{c}{\textbf{0.1}}&&\multicolumn{2}{c}{\textbf{0.2}}&& \multicolumn{2}{c}{\textbf{0.4}} && \multicolumn{2}{c}{\textbf{0.6}} && \multicolumn{2}{c}{\textbf{0.8}} && \multicolumn{2}{c}{\textbf{1}} \\
    & $\bm{r_p}$\ &\ $\bm{r_s}$&&$\bm{r_p}$\ &\ $\bm{r_s}$&&$\bm{r_p}$\ &\ $\bm{r_s}$&&$\bm{r_p}$\ &\ $\bm{r_s}$&&$\bm{r_p}$\ &\ $\bm{r_s}$&& $\bm{r_p}$\ &\ $\bm{r_s}$ \\   
    \midrule
    Human&.965&.923&&.981&.954&&.990&.968&&.992&.972&&\textbf{.994}&.972&&.994&.976\\
    \textsc{PEEM}&\textbf{.978}&\textbf{.962}&&\textbf{.986}&\textbf{.975}&&\textbf{.991}&\textbf{.982}&&\textbf{.993}&\textbf{.985}&&.993&\textbf{.987}&&\textbf{.994}&\textbf{.986}\\
    \bottomrule
    \end{tabular}
    \end{sc}
    \end{small}
    \end{center}
\end{table*}

\paragraph{Impact of Varying Sample Size.} 
With the increase of $q_{sample}$, all algorithms show improved performance accordingly as $Condition\ 1$ is satisfied to a greater extent. We also notice that $r_s$ and $r_p$ exhibit logarithmic curves as $q_{sample}$ increases (see Figure~\ref{fig:acc_pea_sample}). 
This inspires us that the benefits of increasing sample size to attain more accurate evaluation results are greatest in the early stages. Meanwhile, this also demonstrates that when the number of samples is sufficiently large, the impact on evaluation accuracy of not meeting Condition 1 can be minimal.

\subsection{Classification Task}
\label{exp3}
To further validate \textsc{PoEM} under scenarios where inaccurate labels are available, we experiment within the context of classification task on MultiRC \citep{multirc} benchmark.

\subsubsection{Experimental Setup}
MultiRC evaluates the reading comprehension ability of models by having them answer multiple-choice questions. We randomly select 200 samples to conduct experiments for saving inference costs. We use the macro-average $F1$ score to calculate $Cons(Y,\hat{Y}^i)$ following \citet{multirc}:
\begin{equation}
    \small
    \label{ex:eq3}
    Cons(Y,\hat{Y}^i) = \frac{1}{NT(Y)T(\hat{Y}^i)}\sum_{j=1}^N\sum_{u=1}^{T(Y)}\sum_{v=1}^{T(\hat{Y}^i)}F1(y_{j,u},\hat{y}_{j,v}^i)
\end{equation}
Each sample in MultiRC is annotated by multiple well-trained annotators, and the ensemble result of multiple annotations is considered the true label by \citet{multirc}. The annotations made by a certain person can be regarded as the human-level labels $\hat{y}^{human}$, which contain inaccurate ones. Traditionally, $Cons(\hat{y}^{human},\hat{y}^i)$ is used to measure the capability of $\mathcal{M}^i$ if only one annotator is available. 
Considering human as a model under evaluation, we take the consistency matrix among $\{\hat{y}^{i}\}_{i \in \{human\} \cup \{1,...,L\}}$ as inputs to verify whether \textsc{PoEM} can assist humans in better evaluating model capabilities.

\subsubsection{Experimental Results}
As shown in Table~\ref{tab:exp4}, we find that \textsc{PoEM} surpasses traditional human-centric evaluation approach by an obvious margin under poor supervision, especially when sample size is small. From this perspective, \textsc{PoEM} has advanced the evolution of evaluation paradigm from human-centric to human$\And$model-centric paradigm, using mutual consistency across both humans and models to compensate for the insufficiencies of human evaluating abilities in the era of LLMs.

\section{Conclusions}
In this paper, we propose \textsc{PoEM}, a theoretically grounded framework for conducting model evaluation under poor supervision (without true labels). 
Based on EM algorithm, \textsc{PoEM} alternately optimizes model weights and filters reference models to better assess model abilities through their mutual consistency.
Comprehensive experiments validate the effectiveness, efficiency and generalizability of \textsc{PoEM} across multiple tasks and LLMs. 

\textbf{Limitation.} 
Condition 1 and Condition 3 of Theorem 1 are strong assumptions in real-world scenarios, with Condition 1 (infinite sample size) being even impossible. 
Fortunately, based on the experimental results, $\mathcal{F}_{\textsc{PoEM}}$ has effectively alleviated such issues, thereby achieving high consistency with supervised evaluation. 
We look forward to future research that can explore the relationship between the degree of condition fulfillment and the evaluation results, and further optimize $\mathcal{F}_{\textsc{PoEM}}$.

\section*{Ethics Statement}
All of the datasets used in this study were publicly available, and no annotators were employed for our data collection. We confirm that the datasets we used did not contain any harmful content and was consistent with their intended use (research). We have cited the datasets and relevant works used in this study.
\section*{Acknowledgments}
This work is supported by the Beijing Natural
Science Foundation, China (Nos. 4222037, L181010).

\bibliography{acl_latex}
\clearpage
\appendix
\clearpage
\section{Proof of Theorem~\ref{theorem1}}
\label{theorem_proof}
\newtheorem{definition}{Definition}
\begin{definition}
\label{def:inj}
Given models $M^i,M^j$ under evaluation, reference model $\dot{M}$, benchmark $\mathbb{D}: (X=\{x_i\}^N_{i=1},Y=\{y_i\}^N_{i=1})$ and predictions of $M^i$ on $X$ as $\hat{Y}^i=\{\hat{y}^i_j\}^N_{j=1}$, the consistency between Model $M^i$ and Ground Truth $Y$, denoted as $Cons(\hat{Y}^i,Y)$, is typically used to measure the performance of Model $M^i$.
We prove that the following equation holds:
\begin{equation}
\setlength{\jot}{2pt}
\small
\begin{gathered}
Cons(\dot{Y},\hat{Y}^u)<Cons(\dot{Y},\hat{Y}^v)\\
\Leftrightarrow Cons(Y,\hat{Y}^u)<Cons(Y,\hat{Y}^v)
\end{gathered}
\label{equation: appendix}
\end{equation}
when condition 1: Both $P(\dot{Y}|X), P(\hat{Y}^i|X)$ and $P(\dot{Y}|X), P(\hat{Y}^j|X)$ are independently distributed; condition 2: $\dot{M}$ performs better than random predictor; condition 3: $N \to \infty$.
\end{definition}

\begin{proof} 
We prove the theorem separately in the continuous and discrete label domains as follows:

\paragraph{Discrete Domain}
In discrete label domain, we define $Cons(A,B) = Avg(\mathbf{1}_{a_i=b_i}|a_i\in A,b_i\in B)$. Without loss of generality, we assume that the label domain has $T$ possible values and the distribution of $P(\hat{Y}^i|X)$ are as follows:
\begin{equation}
\label{eq:1}
\small
P(\hat{y}_j^i=t|x_j) = \begin{cases}
\sigma_j^i, \ t= y_j
 \\
(1-\sigma_j^i)\times \lambda_j^{i,t},\ t \neq y_j\  (\sum_{t \neq y_j} \lambda_j^{i,t} = 1)
\end{cases}
\end{equation}
As $\dot{M}$ performs better than random predictor, $\mathbb{E}(\dot{\sigma})>\frac{1}{T}$ holds. Based on the assumptions above, we can derive as follows:
\begin{equation}
\begin{split}
\label{eq:2}
&\ \ \ Cons(\hat{Y}^u,\dot{Y}) < Cons(\hat{Y}^v,\dot{Y}) \\
&\Leftrightarrow \sum^N_{i=1} Cons(\hat{y}_i^u,\dot{y}_i)<\sum^N_{i=1} Cons(\hat{y}_i^v,\dot{y}_i) \\
&\Leftrightarrow \sum^N_{i=1} \mathbf{1}_{\hat{y}_i^u=\dot{y}_i} < \sum^N_{i=1} \mathbf{1}_{\hat{y}_i^v=\dot{y}_i} \\
&\Leftrightarrow \sum^N_{i=1} \sigma_i^u < \sum^N_{i=1} \sigma_i^v \\
&\Leftrightarrow N\mathbb{E}(\sigma^u) < N\mathbb{E}(\sigma^v)\ \ \ ,according\ to\ condition\ 3\\
&\Leftrightarrow \mathbb{E}(\sigma^u) < \mathbb{E}(\sigma^v)
\end{split}
\end{equation}
From another direction:
\begin{equation}
\begin{split}
\label{eq:3}
&\ \ \ Cons(\hat{Y}^u,\dot{Y}) < Cons(\hat{Y}^v,\dot{Y}) \\
&\Leftrightarrow \sum^N_{i=1} Cons(\hat{y}_i^u,\dot{y}_i)<\sum^N_{i=1} Cons(\hat{y}_i^v,\dot{y}_i) \\
&\Leftrightarrow \sum^N_{i=1} \mathbf{1}_{\hat{y}_i^u=\dot{y}_i} < \sum^N_{i=1} \mathbf{1}_{\hat{y}_i^v=\dot{y}_i} \\
&\Leftrightarrow \sum^N_{i=1} (\sigma_i^u\dot{\sigma}_i+\sum_{t\neq y_i} \lambda_i^{u,t}\dot{\lambda}_i^{t})< \sum^N_{i=1} (\sigma_i^v\dot{\sigma}_i+\sum_{t\neq y_i} \lambda_i^{v,t}\dot{\lambda}_i^{t}) \\
&\Leftrightarrow N\mathbb{E}(\sigma^u\dot{\sigma}+\sum_{t\neq y_i} \lambda_i^{u,t}\dot{\lambda}_i^{t})< N\mathbb{E}(\sigma^v\dot{\sigma}+\sum_{t\neq y_i} \lambda_i^{v,t}\dot{\lambda}_i^{t})\ \ \ \\
&\ \ \ \ \ \ ,according\ to\ condition\ 3 \\
&\Leftrightarrow \mathbb{E}(\sigma^u)\mathbb{E}(\dot{\sigma})+(T-1)\mathbb{E}(\lambda^{u})\mathbb{E}(\dot{\lambda}) \\
&\ \ \ \ \ \ < \mathbb{E}(\sigma^v)\mathbb{E}(\dot{\sigma})+(T-1)\mathbb{E}(\lambda^{v})\mathbb{E}(\dot{\lambda})\ \ \ \\
&\ \ \ \ \ \ ,according\ to\ condition\ 1\\
&\Leftrightarrow \mathbb{E}(\sigma^u)\mathbb{E}(\dot{\sigma})+(1-\mathbb{E}(\sigma^u))(1-\mathbb{E}(\dot{\sigma}))/(T-1)\\
&\ \ \ \ \ \ <\mathbb{E}(\sigma^v)\mathbb{E}(\dot{\sigma})+(1-\mathbb{E}(\sigma^v))(1-\mathbb{E}(\dot{\sigma}))/(T-1) \\
&\Leftrightarrow (\mathbb{E}(\sigma^u)-\mathbb{E}(\sigma^v))(\mathbb{E}(\dot{\sigma})-\frac{1-\mathbb{E}(\dot{\sigma})}{T-1})<0 \\
&\Leftrightarrow (\mathbb{E}(\sigma^u)-\mathbb{E}(\sigma^v))(T\mathbb{E}(\dot{\sigma})-1)<0 \\
&\Leftrightarrow \mathbb{E}(\sigma^u)<\mathbb{E}(\sigma^v)\ \ \ ,according\ to \ condition\ 2
\end{split}
\end{equation}
According to Eq~\ref{eq:2} and Eq~\ref{eq:3}, Eq~\ref{equation: appendix} holds in discrete label domain.

\paragraph{Continuous Domain}
In continuous label domain, we define $Cons(A,B) = -Avg(Abs(a_i-b_i)|a_i\in A,b_i\in B)$. Without loss of generality, we assume $\hat{
y}^u_i=y_i+g_i^u$, where $g_i^u$ is zero-mean Gaussian noise with variance as $\sigma_i^u$. Based on the assumptions above, we can derive as follows:
\begin{equation}
\begin{split}
\label{eq:5}
&\ \ \ Cons(\hat{Y}^u,\dot{Y}) < Cons(\hat{Y}^v,\dot{Y}) \\
&\Leftrightarrow \sum^N_{i=1} Cons(\hat{y}_i^u,\dot{y}_i)<\sum^N_{i=1} Cons(\hat{y}_i^v,\dot{y}_i) \\
&\Leftrightarrow \sum^N_{i=1} |\hat{y}_i^u-\dot{y}_i| > \sum^N_{i=1} |\hat{y}_i^v-\dot{y}_i| \\
&\Leftrightarrow \sum^N_{i=1} |g_i^u| > \sum^N_{i=1} |g_i^v| \\
&\Leftrightarrow N\mathbb{E}(|g^u|) > N\mathbb{E}(|g^v|)\ \ \ ,according\ to\ condition\ 3\\
&\Leftrightarrow \mathbb{E}(\sigma^u) > \mathbb{E}(\sigma^v)
\end{split}
\end{equation}
From another direction:
\begin{equation}
\begin{split}
\label{eq:6}
&\ \ \ Cons(\hat{Y}^u,\dot{Y}) < Cons(\hat{Y}^v,\dot{Y}) \\
&\Leftrightarrow \sum^N_{i=1} Cons(\hat{y}_i^u,\dot{y}_i)<\sum^N_{i=1} Cons(\hat{y}_i^v,\dot{y}_i) \\
&\Leftrightarrow \sum^N_{i=1} |\hat{y}_i^u-\dot{y}_i| > \sum^N_{i=1} |\hat{y}_i^v-\dot{y}_i| \\
&\Leftrightarrow N\mathbb{E}(|\hat{y}^u-\dot{y}|)>N\mathbb{E}(|\hat{y}^v-\dot{y}|)\ \ \ \\
&\ \ \ \ \ \ ,according\ to\ condition\ 1 \\
&\Leftrightarrow N*(\mathbb{E}(\sigma^u)+\mathbb{E}(\dot{\sigma}))> N*(\mathbb{E}(\sigma^v)+\mathbb{E}(\dot{\sigma}))\ \ \ \\
&\ \ \ \ \ \ ,according\ to\ condition\ 2 \\
&\Leftrightarrow \mathbb{E}(\sigma^u)>\mathbb{E}(\sigma^v)
\end{split}
\end{equation}
According to Eq~\ref{eq:5} and Eq~\ref{eq:6}, Eq~\ref{equation: appendix} holds in discrete label domain.
\end{proof}

\section{Detailed Statistics}
We conduct experiments with 16 mainstream LLMs as shown in Table~\ref{tab:exp0}. The detailed statistics of MATH benchmark are shown in Table~\ref{tab:exp2}.

\section{Further Experimental Results}

\subsection{Results with Greedy Search}
\label{sec:greedy}
Apart from the main results of the proposed algorithms with random sampling at temperature as 0.5 and sampling times as 5, we also show the results with greedy search as shown in Table~\ref{tab:greedy}. $\mathcal{F}_{\textsc{PoEM}}$ performs best as expected. Compared with random sampling at sampling times as 5, the correlations between $\hat{B}$ and $B$ decrease a little bit across all the algorithms. We believe this is because multiple sampling can eliminate the noise impact brought by single sampling.

\subsection{Hyperparameter Analysis}
\label{sec:p}
We have explored the effectiveness of $\mathcal{F}_{filter}$ under different values of $p$. As shown in Table~\ref{tab:p}, $\mathcal{F}_{filter}$ performs relatively stably across different p-values, reaching its best when $p=0.9$.

\section{Applicability of \textsc{PoEM} in Open-Domain Tasks}
The experiments above have verified the effectiveness of \textsc{PoEM} in various tasks within closed label domain. Meanwhile, open-domain text generation \citep{ret0,ret1} is also an important aspect of LLM capability assessment. Unlike closed domain tasks holding a one-to-one relationship between input and output, open-domain tasks can have multiple appropriate answers for a given input. Therefore, evaluation based on consistency may not be suitable for open-domain evaluation and current mainstream methods rely on human-evaluator and model-evaluator \citep{geval,eva1,eva2}.
However, we still need a method to choose the most capable evaluator. The output of certain evaluator is closed-domain, where we can use \textsc{PoEM} to carry out such evaluations, as demonstrated in \S\ref{exp2}. To summarize, \textsc{PoEM} can enhance the accuracy of model capability assessment in open-domain tasks by aiding in the selection of evaluators.

\section{Links of Assets Involved}
\textbf{LLMs weights:}
\begin{enumerate}
\item BAICHUAN-7B:\url{https://huggingface.co/baichuan-inc/Baichuan-7B}
\item VICUNA-7B-V1.5:\url{https://huggingface.co/lmsys/vicuna-7b-v1.5}
\item LLAMA-2-7B-CHAT:\url{https://huggingface.co/meta-llama/Llama-2-7b-chat}
\item MISTRAL-7B-INSTRUCT-V0.2:\url{https://huggingface.co/mistralai/Mistral-7B-Instruct-v0.2}
\item METAMATH-7B-V1.0:\url{https://huggingface.co/meta-math/MetaMath-7B-V1.0}
\item WIZARDMATH-7B-V1.1:\url{https://huggingface.co/WizardLMTeam/WizardMath-7B-V1.1}
\item OPENCHAT-3.5-0106:\url{https://huggingface.co/openchat/openchat-3.5-0106}
\item VICUNA-13B-V1.5:\url{https://huggingface.co/lmsys/vicuna-13b-v1.5}
\item LLAMA-2-13B-CHAT-HF:\url{https://huggingface.co/meta-llama/Llama-2-13b-chat-hf}
\item METAMATH-13B-V1.0:\url{https://huggingface.co/meta-math/MetaMath-13B-V1.0}
\item LLAMA-2-70B-CHAT-HF:\url{https://huggingface.co/meta-llama/Llama-2-70b-chat-hf}
\item METAMATH-70B-V1.0:\url{https://huggingface.co/meta-math/MetaMath-70B-V1.0}
\end{enumerate}

\textbf{API documents:}
\begin{enumerate}
\item GEMINI-PRO:\url{https://ai.google.dev/}
\item GPT-series:\url{https://openai.com/index/openai-api/}
\end{enumerate}

\textbf{Benchmarks:}
\begin{enumerate}
\item MATH:\url{https://people.eecs.berkeley.edu/~hendrycks/MATH.tar}
\item USR:\url{http://shikib.com/tc_usr_data.json}
\item MultiRC:\url{https://cogcomp.seas.upenn.edu/multirc/data/mutlirc-v2.zip}
\end{enumerate}

\begin{table*}[ht]
    \caption{Model-level Pearson ($r_p$) / Spearman ($r_s$) correlations of different algorithms on 7 subsets of MATH with greedy search. P-values of all the results < 0.05.}\label{tab:greedy}
    \begin{center}
    \begin{small}
    \begin{sc}
    \setlength{\tabcolsep}{0.13em} 
    \begin{tabular}{lccp{0.005pt}ccp{0.005pt}ccp{0.005pt}ccp{0.005pt}ccp{0.005pt}ccp{0.005pt}ccp{0.005pt}cc}
    \toprule
    \multirow{2}{*}{\textbf{Subset}} &\multicolumn{2}{c}{\textbf{IA}}&& \multicolumn{2}{c}{\textbf{PA}} && \multicolumn{2}{c}{\textbf{GE}} && \multicolumn{2}{c}{\textbf{CP}} && \multicolumn{2}{c}{\textbf{AG}}&& \multicolumn{2}{c}{\textbf{NT}}&& \multicolumn{2}{c}{\textbf{PC}}&& \multicolumn{2}{c}{\textbf{Average}} \\
    & $\bm{r_p}$\ &\ $\bm{r_s}$&&$\bm{r_p}$\ &\ $\bm{r_s}$&&$\bm{r_p}$\ &\ $\bm{r_s}$&&$\bm{r_p}$\ &\ $\bm{r_s}$&&$\bm{r_p}$\ &\ $\bm{r_s}$&& $\bm{r_p}$\ &\ $\bm{r_s}$&& $\bm{r_p}$\ &\ $\bm{r_s}$&& $\bm{r_p}$\ &\ $\bm{r_s}$ \\   
    \midrule
    Ensemble&.679&.818&&.956&.993&&.910&.960&&.856&.966&&.949&.994&&.857&.969&&.773&.867&&.854&.938\\
    Calibration&.754&.913&&.938&.994&&.893&.986&&.835&.980&&.922&.993&&.835&.974&&.827&.953&&.858&.971\\
    Filtering&.672&.881&&.982&.996&&.931&.988&&.932&.985&&.970&.998&&.923&.974&&.796&.930&&.886&.965\\
    PoEM&.779&.947&&.976&.995&&.935&.992&&.931&.986&&.971&.997&&.924&.975&&.751&.962&&\textbf{.895}&\textbf{.979}\\
    \bottomrule
    \end{tabular}
    \end{sc}
    \end{small}
    \end{center}
\end{table*}

\begin{table*}[ht]
    \vskip 0.15in
    \caption{Model-level Pearson ($r_p$) / Spearman ($r_s$) correlations of $\mathcal{F}_{filter}$ on 7 subsets of MATH with different values of $p$. P-values of all the results < 0.05.}\label{tab:p}
    \begin{center}
    \begin{small}
    \begin{sc}
    \setlength{\tabcolsep}{0.09em} 
    \begin{tabular}{lccp{0.005pt}ccp{0.005pt}ccp{0.005pt}ccp{0.005pt}ccp{0.005pt}ccp{0.005pt}ccp{0.005pt}cc}
    \toprule
    \multirow{2}{*}{\textbf{Subset}} &\multicolumn{2}{c}{\textbf{IA}}&& \multicolumn{2}{c}{\textbf{PA}} && \multicolumn{2}{c}{\textbf{GE}} && \multicolumn{2}{c}{\textbf{CP}} && \multicolumn{2}{c}{\textbf{AG}}&& \multicolumn{2}{c}{\textbf{NT}}&& \multicolumn{2}{c}{\textbf{PC}}&& \multicolumn{2}{c}{\textbf{Average}} \\
    & $\bm{r_p}$\ &\ $\bm{r_s}$&&$\bm{r_p}$\ &\ $\bm{r_s}$&&$\bm{r_p}$\ &\ $\bm{r_s}$&&$\bm{r_p}$\ &\ $\bm{r_s}$&&$\bm{r_p}$\ &\ $\bm{r_s}$&& $\bm{r_p}$\ &\ $\bm{r_s}$&& $\bm{r_p}$\ &\ $\bm{r_s}$&& $\bm{r_p}$\ &\ $\bm{r_s}$ \\   
    \midrule
    Filtering $p=0.8$&.831&.892&&.987&.995&&.966&.990&&.951&.994&&.983&.998&&.943&.987&&.886&.952&&.935&\textbf{.972}\\
    Filtering $p=0.9$&.834&.886&&.997&.998&&.986&.990&&.980&.994&&.995&.999&&.969&.983&&.889&.931&&\textbf{.950}&.969\\
    Filtering $p=1.0$&.786&.917&&.995&.998&&.972&.991&&.964&.989&&.995&.999&&.945&.981&&.815&.932&&.925&\textbf{.972}\\
    \bottomrule
    \end{tabular}
    \end{sc}
    \end{small}
    \end{center}
\end{table*}

\begin{table}[t]
\vskip 0.15in
\caption{Statistics of LLMs involved in the experiments.}\label{tab:exp0}
\begin{center}
\begin{small}
\begin{sc}
\begin{tabular}{lcc}
\toprule
Model & Size (B)&Inference \\
\midrule
Baichuan-7B    & 7 &A100 40G\\
vicuna-7b-V1.5 & 7&A100 40G\\
llama-2-7b-chat    & 7 &A100 40G\\
Mistral-7B-Instruct-v0.2    & 7&A100 40G         \\
MetaMath-7B-V1.0     & 7&A100 40G\\
WizardMath-7B-V1.1      & 7&A100 40G \\
openchat-3.5-0106      & 7&A100 40G  \\
vicuna-13b-V1.5&13&A100 40G \\
Llama-2-13b-chat-hf &13&A100 40G\\
MetaMath-13B-V1.0&13&A100 40G\\
Llama-2-70b-chat-hf&70&A100 40G\\
MetaMath-70B-V1.0&70&A100 40G\\
Gemini-pro&-&API\\
gpt-3.5-turbo (0613)&-&API\\
gpt-4 (0613)&-&API\\
gpt-4-turbo (1106)&-&API\\
\bottomrule
\end{tabular}
\end{sc}
\end{small}
\end{center}
\vskip -0.1in
\end{table}

\begin{table}[t]
\vskip 0.15in
\caption{Statistics of MATH's subsets.}\label{tab:exp2}
\begin{center}
\begin{small}
\begin{sc}
\begin{tabular}{lcc}
\toprule
Subset & Abbreviation&Num \\
\midrule
Intermediate Algebra & IA &903\\
Prealgebra & PA&871\\
Geometry & GE &479\\
Counting \& Probability & CP&474 \\
Algebra & AG&1187\\
Number Theory & NT&540 \\
Precalculus &PC & 546  \\
\bottomrule
\end{tabular}
\end{sc}
\end{small}
\end{center}
\vskip -0.1in
\end{table}

\end{document}